\pdfoutput=1

\documentclass[11pt]{article}

\usepackage{acl}

\usepackage{times}
\usepackage{latexsym}

\usepackage[T1]{fontenc}

\usepackage[utf8]{inputenc}

\usepackage{microtype}

\usepackage{inconsolata}

\usepackage{amsmath}
\usepackage{booktabs} %
\usepackage{multirow}
\usepackage{graphicx}
\usepackage{xurl}
\usepackage{subcaption}
\usepackage{amssymb}

\title{MSLM-S2ST: A Multitask Speech Language Model for Textless Speech-to-Speech Translation with Speaker Style Preservation}

\author{Yifan Peng$^1$\thanks{\quad Work done during an internship at Meta AI.} , Ilia Kulikov$^2$, Yilin Yang$^2$, Sravya Popuri$^2$, Hui Lu$^{3*}$, \\{\bf Changhan Wang$^2$, Hongyu Gong$^2$} \\
$^1$ Carnegie Mellon University\quad
$^2$ Meta AI \quad
$^3$ The Chinese University of Hong Kong\\
}

\begin{document}
\maketitle
\begin{abstract}
There have been emerging research interest and advances in speech-to-speech translation (S2ST), translating utterances from one language to another. This work proposes \textbf{M}ultitask \textbf{S}peech \textbf{L}anguage \textbf{M}odel (MSLM), which is a decoder-only speech language model trained in a multitask setting. Without reliance on text training data, our model is able to support \textbf{multilingual} S2ST with speaker style preserved. 
\end{abstract}

\section{Introduction}

The study of speech-to-speech translation (S2ST) is motivated by the communication need of speakers talking in different languages. 
Most existing S2ST systems employ an encoder-decoder architecture, and Translatotron 2~\citep{translatotron2} uses text as intermediate outputs when decoding spectrograms. Research progress made in textless NLP aims to remove dependency on text, which is generalizable to languages without a writing system.
Textless S2ST is one of these efforts~\citep{lee-etal-2022-textless}.

Inspired by the potential of language models (LMs) in capturing contextual information and long-range dependency, there is a surge of research interest in speech language models~\citep{lakhotia-etal-2021-generative, kharitonov-etal-2022-text, nguyen-etal-2023-generative, audiolm, Speechlmscore, valle, vallex, polyvoice, s2st-style, voxtlm}. 
The challenge of modeling and synthesizing speech lies in the processing of continuous waveforms and acoustic features. Recent advances in speech tokenizers pave the way towards speech language modeling. Self-supervised learning (SSL) based models such as wav2vec 2.0~\citep{wav2vec2} and HuBERT~\citep{hubert} are shown to learn semantic units from speech, and recent works on neural codec models such as EnCodec~\citep{encodec} and SoundStream~\citep{soundstream} extract acoustic tokens that are suitable for high-quality audio reconstruction.
The GSLM family~\citep{lakhotia-etal-2021-generative, kharitonov-etal-2022-text, nguyen-etal-2023-generative} train autoregressive LMs on HuBERT units. AudioLM~\citep{audiolm} and the VALL-E family~\citep{valle, vallex} model the conditional generation of acoustic tokens and achieve excellent performance on conditional speech synthesis.

\begingroup
\setlength{\tabcolsep}{2pt}
\begin{table*}[tbp!]
\centering
\resizebox {\linewidth} {!} {
\begin{tabular}{lcccc}
\toprule
 & Enc-Dec S2ST & PolyVoice & StyleLM & MSLM-S2ST\\
 & \citep{duquenne-etal-2023-speechmatrix} & \citep{polyvoice} & \citep{s2st-style} & (ours) \\
\midrule\midrule
Style preservation & - & $\checkmark$ & $\checkmark$ & $\checkmark$ \\\midrule
\begin{tabular}[c]{@{}l@{}} Single model for multiple \\ translation directions \end{tabular} & - & $\checkmark$ & - & $\checkmark$ \\\midrule
Semantic unit translation & Encoder-decoder & \begin{tabular}[c]{@{}l@{}} An AR LM for unit translation,\\ an AR LM for duration prediction \end{tabular} & Encoder-decoder & AR LM (shared) \\\midrule
Acoustic unit generation & - & AR LM and NAR LM & AR LM & AR LM (shared) and NAR LM \\\midrule
Waveform synthesis & Unit vocoder & SoundStream decoder & Unit vocoder & EnCodec decoder \\
\bottomrule
\end{tabular}
}
\caption{Comparison of our MSLM with recent S2ST methods. Our MSLM improves the efficiency of style-preserved S2ST systems by supporting multiple translation directions in a single model and sharing the AR LM for semantic-to-semantic translation and semantic-to-acoustic generation.
}
\vskip -0.1in
\label{tab:mslm-vs-prior-s2st}
\end{table*}
\endgroup

Inspired by these studies, we propose a \textbf{M}ultitask \textbf{S}peech \textbf{L}anguage \textbf{M}odel (MSLM) for textless S2ST with speaker style preservation. MSLM leverages semantic units from HuBERT and acoustic units from EnCodec. We formulate a multitask learning framework using a decoder-only autoregressive (AR) LM, which performs two tasks in a single model: (1) semantic-to-semantic translation, where the AR LM translates semantic units from the source language into the target language; (2) semantic-to-acoustic generation, where the AR LM generates target acoustic units conditioned on target semantic units and source speech as the style prompt. Since acoustic units have multiple streams, the AR LM takes care of the first stream, and another non-autoregressive (NAR) LM is utilized to predict the remaining streams.

We highlight contributions of the proposed approach as follows:

\noindent(1) Compared to S2ST systems relying on text data \cite{vallex,audiopalm}, our approach is speech-only training, which does not need any text. Therefore, it can be applied to unwritten languages.

\noindent(2) We propose a single speech LM that supports speaker style-preserved S2ST. The most relevant work is PolyVoice~\citep{polyvoice} in this direction, but we note that it involves three separate LMs for semantic unit translation, duration prediction and acoustic unit generation, respectively.~\footnote{Here ``acoustic unit generation'' contains only the first EnCodec stream. To generate the remaining streams, both our work and PolyVoice utilize a secondary NAR LM.} %

\noindent(3) Our model can achieve multilingual S2ST 
    and demonstrates high translation quality and speaker style similarity between English and Spanish, while most S2ST works are built for a specific translation direction including PolyVoice~\citep{polyvoice}.

\section{Related Work}
\label{sec:related-work}

With recent advances in discrete audio representations \citep{wav2vec2,hubert}, speech LMs have attracted an increasing research interest such as GSLM family \citep{lakhotia-etal-2021-generative, kharitonov-etal-2022-text, nguyen-etal-2023-generative}, VoxtLM \citep{voxtlm}, VALL-E \citep{valle} and AudioPaLM \citep{audiopalm}. Recent studies have explored speech LMs for S2ST, employing separate translation and acoustic unit generation \cite{polyvoice, audiopalm, s2st-style}. This work simplifies the overall architecture and improves efficiency. Please refer to \autoref{app:related-work} for more discussions.

\begin{figure}[tb!]
     \centering
     \begin{subfigure}[t]{\linewidth}
         \centering
         \includegraphics[scale=0.51]{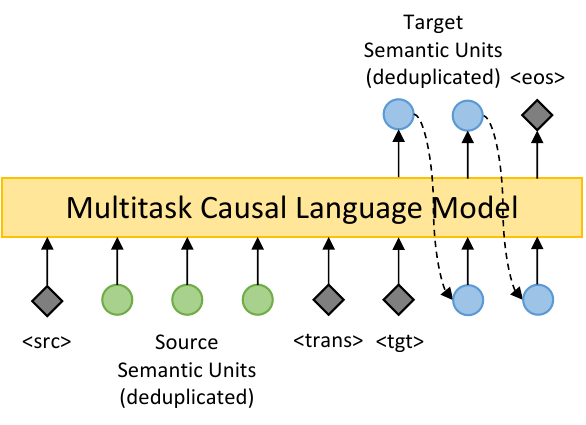}
         \vskip -0.07in
         \caption{Our MSLM translates semantic units from the source language to the target language. \texttt{<src>} and \texttt{<tgt>} are language tags. \texttt{<trans>} denotes the ``translation'' task. \texttt{<eos>} is the end-of-sentence token.}
         \label{fig:semantic2semantic}
     \end{subfigure}
     \begin{subfigure}[t]{\linewidth}
         \centering
         \includegraphics[scale=0.51]{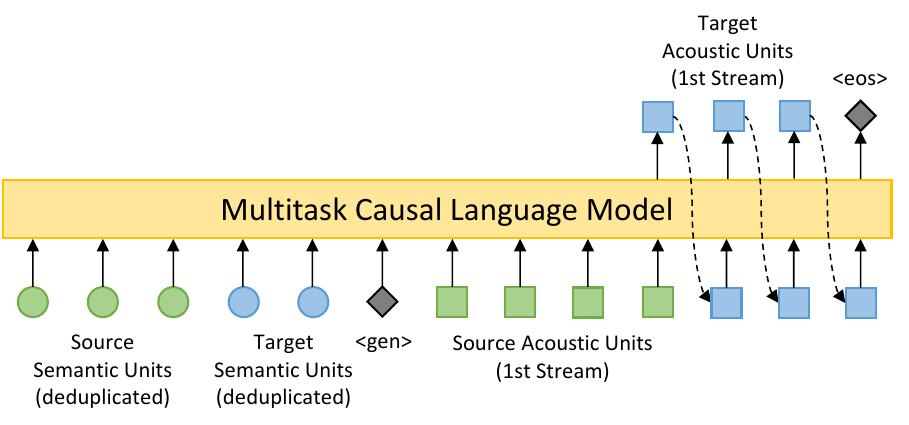}
         \vskip -0.07in
         \caption{Our MSLM generates the first stream of acoustic units given the translated semantic units and the style prompt. \texttt{<gen>} denotes the ``generation'' task.}
         \label{fig:semantic2acoustic1}
     \end{subfigure}
     \begin{subfigure}[t]{\linewidth}
         \centering
         \includegraphics[scale=0.51]{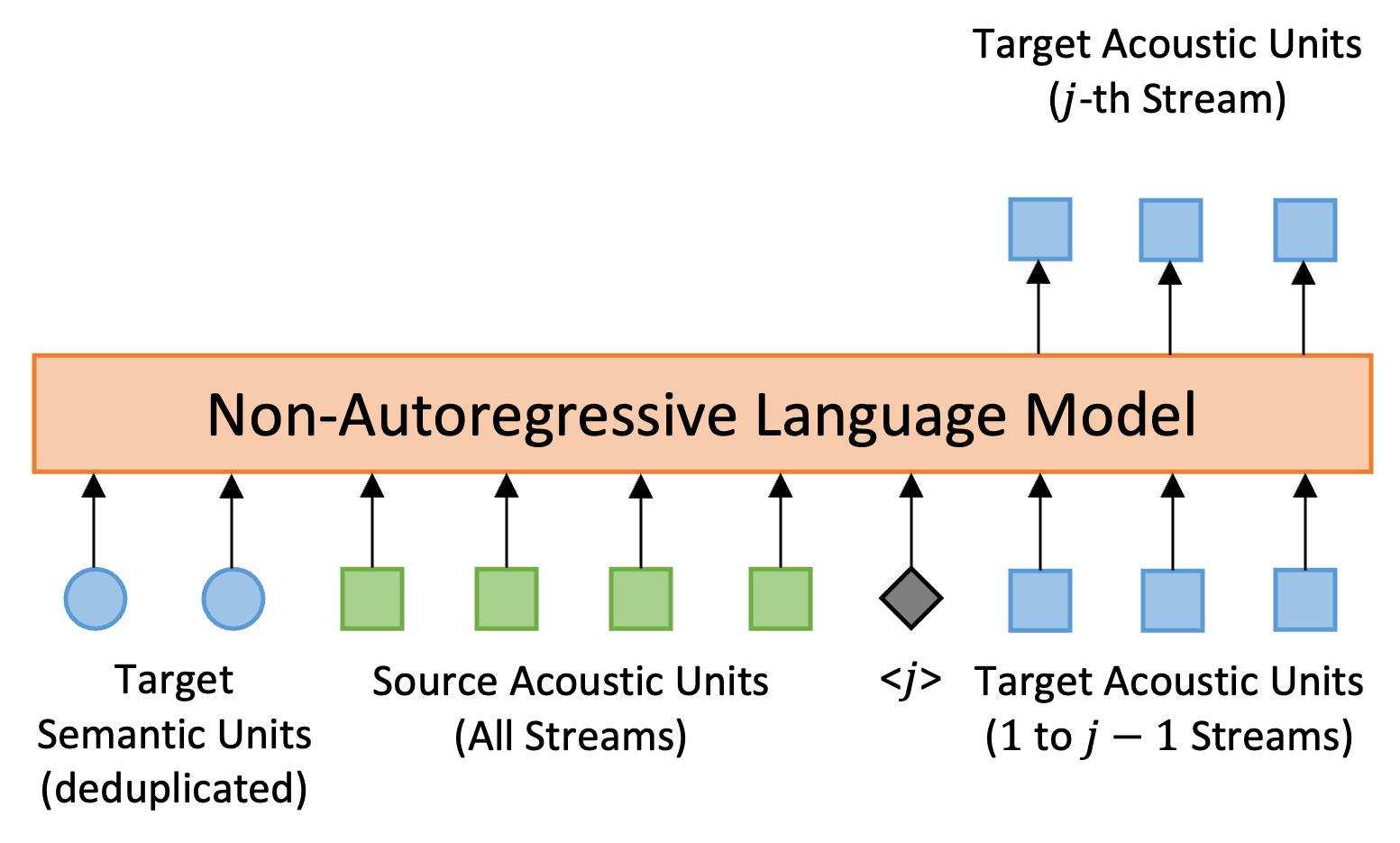}
         \vskip -0.07in
         \caption{Another non-autoregressive LM generates the remaining streams of acoustic units. \texttt{<j>} indicates the stream to be generated.}
         \label{fig:semantic2acoustic2}
     \end{subfigure}
    \caption{Inference procedure of speaker style-preserved S2ST. Previous studies~\cite{polyvoice, s2st-style} typically use separate LMs for each step and each translation direction, whereas our MSLM performs (a) and (b) in a single AR LM controlled by a special task token. MSLM also supports multilingual translation controlled by special language tokens.}
    \label{fig:full-procedure}
    \vskip -0.1in
\end{figure}

\section{Proposed Method}
\label{sec:proposed-method}

We extract semantic units from the source speech using a pre-trained HuBERT and translate them into the target language. Then, we generate multiple streams of acoustic units conditioned on the predicted target semantic units and the source acoustic units as a style prompt. Finally, we synthesize target waveforms using the pre-trained EnCodec decoder. \autoref{app:method} illustrates this procedure.

\subsection{Speech unit extraction}

We use HuBERT for speech semantic units \citep{hubert,s2st-multi-target}, and EnCodec for multi-stream acoustic units \citep{encodec}.

\subsection{Semantic-to-semantic translation}

As shown in \autoref{fig:semantic2semantic}, given a sequence of semantic units in the source language $S^{\mathrm{src}}$, our decoder-only multitask speech LM (MSLM) predicts a sequence of semantic units in the target language $S^{\mathrm{tgt}}$ in an autoregressive manner:
\begin{align}
\label{eq:mslm-semantic-semantic}
\begin{split}
    &\hat{S}^{\mathrm{tgt}}_{i} \sim \\
    &p_{\mathrm{MSLM}}(S^{\mathrm{tgt}}_{i} \mid \texttt{<src>}, S^{\mathrm{src}}, \texttt{<trans>}, \texttt{<tgt>}, S^{\mathrm{tgt}}_{<i}),
\end{split}
\end{align}
where \texttt{<src>} and \texttt{<tgt>} are special tokens denoting the source and target languages, respectively. \texttt{<trans>} is another special token specifying the ``translation'' task.

\subsection{Semantic-to-acoustic generation}

Following the design of VALL-E~\citep{valle, vallex}, the 8 streams of acoustic units are generated in two sequential steps. The first step generates only the first stream using our autoregressive MSLM (\autoref{fig:semantic2acoustic1}), while the second step generates the other streams progressively using another non-autoregressive LM (\autoref{fig:semantic2acoustic2}).

Specifically, the first step is defined as follows:
\begin{align}
\label{eq:mslm-semantic-acoustic}
    \hat{A}^{\text{tgt}}_{i,1} \sim p_{\text{MSLM}}(A^{\text{tgt}}_{i,1} \mid S^{\mathrm{src}}, S^{\mathrm{tgt}}, \texttt{<gen>}, A^{\text{src}}_{:,1}, A^{\text{tgt}}_{<i,1}),
\end{align}
where the model MSLM shares the same parameters as that in \autoref{eq:mslm-semantic-semantic}, and \texttt{<gen>} is a special token specifying the ``generation'' task.

The second step is formulated as follows:
\begin{align}
\label{eq:narlm}
    \hat{A}^{\text{tgt}}_{:,j} \sim p_{\text{NARLM}}(A^{\text{tgt}}_{:,j} \mid S^{\mathrm{tgt}}, A^{\text{src}}_{:,1:8}, \texttt{<j>}, A^{\text{tgt}}_{:,1:j-1}),
\end{align}
where $j = 2, \ldots, 8$ is the stream ID. Unlike VALL-E~\citep{valle} using adaptive layer normalization to inject the stream ID, we directly insert a special token \texttt{<j>} in the input sequence. To preserve the style of the source speaker, we condition the generation on source acoustic units $A^{\text{src}}_{:,1:8}$ extracted from up to 5 seconds of audio.

Note that the semantic-to-acoustic generation steps in \autoref{eq:mslm-semantic-acoustic} and \autoref{eq:narlm} are trained on unlabelled speech data.

\subsection{Multitask training}
\label{subsec:multitask-training}

Different from prior studies which employ separate LMs for semantic unit translation and acoustic unit generation~\citep{polyvoice, audiopalm, s2st-style}, we utilize a single decoder-only LM for bidirectional semantic-to-semantic translation (c.f. \autoref{eq:mslm-semantic-semantic}) and semantic-to-acoustic generation (only first stream, c.f. \autoref{eq:mslm-semantic-acoustic}), which reduces the overall model size without performance degradation. Specifically, we mix the training data of both tasks and insert a special task token (\texttt{<trans>} or \texttt{<gen>}) to distinguish between them. The training objective is to minimize the cross-entropy loss for next token prediction, which is a standard target for causal LMs. We empirically find the translation task to be more difficult to learn, so we upsample the translation data by 3 times. We perform various ablation studies in Appendix~\ref{subsec:ablation} to demonstrate the effectiveness of our proposed method.

\begingroup
\begin{table*}[t!]
    \centering
    \resizebox {\linewidth} {!} {
    \begin{tabular}{lcccccccccc}
    \toprule
    \multirow{3}{*}{Model} & & & \multicolumn{4}{c}{Es-En} & \multicolumn{4}{c}{En-Es} \\
    & \multicolumn{2}{c}{\#Parameters} & \multicolumn{2}{c}{Set 1} & \multicolumn{2}{c}{Set 2} & \multicolumn{2}{c}{Set 1} & \multicolumn{2}{c}{Set 2} \\
    & AR & NAR & Spkr Sim. & BLEU & Spkr Sim. & BLEU & Spkr Sim. & BLEU & Spkr Sim. & BLEU \\
    \midrule
    \multicolumn{11}{l}{\textbf{Unidirectional baselines}}\\
    ~~Enc-Dec S2ST + vocoder & \multicolumn{2}{c}{363M} & 0.025 & 26.86 & 0.032 & 24.26 & 0.094  & 21.86 & 0.098 & 21.82 \\
    ~~Cascaded LMs & 306M $\times$ 2 & 306M & 0.391 & 24.24 & 0.380 & 21.51 & 0.439 & 20.98 & 0.430 & 21.82 \\
    ~~AR LM + vocoder & 306M & - & 0.024 & 25.26 & 0.032 & 22.35 & 0.094 & 20.92 & 0.098 & 21.47 \\
    \midrule
    \multicolumn{11}{l}{\textbf{Ours (bidirectional model)}} \\
    ~~MSLM & 306M & 306M & 0.400 & 24.78 & 0.393 & 21.50 & 0.430 & 21.41 & 0.421 & 22.09 \\
    ~~MSLM + vocoder & 306M & - & 0.025 & 24.41 & 0.032 & 21.27 & 0.092 & 20.60 & 0.099 & 21.29 \\
    \bottomrule
    \end{tabular}
}
\caption{Speaker style similarity ($\uparrow$) and ASR-BLEU ($\uparrow$) for Es-En and En-Es S2ST. Each baseline model only supports a single translation direction and thus separate models are needed for each direction, whereas our MSLM supports both directions in a single model.
}
\vskip -0.1in
\label{tab:main-results}
\end{table*}
\endgroup

\section{Experiments}

\subsection{Experimental setup}
\label{subsec:experimental-setup}

\textbf{Speech tokenizers}. We extract HuBERT units as semantic representations of utterances following \citet{lee-etal-2022-textless}. Acoustic units are extracted from an EnCodec model \cite{encodec} trained on 
$24$k-hour English and $21$k-hour Spanish speech
with $8$ codebooks. Both models generate units at a frame rate of $50$Hz. The vocabulary sizes of HuBERT and EnCodec are $1000$ and $1024$. 

\textbf{Training and validation data}. 
We used an in-house dataset which consists of semantically aligned speech in English and Spanish.
There are $500$k samples with 1k-hour parallel speech. We randomly select $5$k samples as the validation set and use the other samples as the training set.

\textbf{Evaluation data}. 
The evaluation data consists of aligned speech-text samples in Spanish-English (Es-En) and English-Spanish (En-Es) directions. We prepared two evaluation sets for each direction, Set 1 has $1947$ sample for Es-En and $1272$ samples for En-Es, and Set 2 has $1816$ and $1267$ samples respectively.

\textbf{Evaluation metrics}. We leverage two metrics to evaluate style-preserved S2ST: (1) ASR-BLEU reflects the translation quality by measuring the BLEU between transcriptions of generated speech and the ground truth target texts. We use Whisper's English ASR model of medium size for English transcriptions\footnote{\url{https://huggingface.co/openai/whisper-medium}}, and XLSR-based Spanish model for Spanish transcriptions\footnote{\url{https://huggingface.co/jonatasgrosman/wav2vec2-large-xlsr-53-spanish}}. (2) Speaker style similarity: following \citet{voicebox}, we make use of a WavLM \cite{wavlm} based speaker style encoder to obtain speaker style embeddings of generated speech and source speech. The cosine similarity between these embeddings is taken as the speaker style similarity.

\textbf{Model setup}. Our S2ST system consists of a causal AR LM and an NAR LM. Both contain $24$ Transformer decoder layers. Each layer has a hidden dimension of $1024$, a feedforward dimension of $4096$ and $16$ attention heads. More model hyperparameters are provided in Appendix~\ref{sec:app_exp}.

\textbf{Baselines}. Two strong baselines are included for empirical comparison.

\noindent(1) \textbf{Enc-Dec S2ST}. We adopt the state-of-the-art S2ST model from SpeechMatrix~\citep{duquenne-etal-2023-speechmatrix}, which is built upon the encoder-decoder architecture. %
With $12$ Transformer encoder layers and $12$ decoder layers, it translates semantic units first and synthesizes speech by a separately trained HiFi-GAN vocoder \cite{polyak21_interspeech}.

\noindent(2) \textbf{Cascaded LMs}. Similar to AudioPaLM \citep{audiopalm} and PolyVoice \citep{polyvoice}, this baseline leverages two \emph{separate} LMs for semantic-to-semantic translation and semantic-to-acoustic generation, respectively. These LMs have the same architecture and are trained with the same hyperparameters as MLSM.

\subsection{Results}

\autoref{tab:main-results} compares MSLM against two baselines for Es<>En S2ST on evaluation sets. Following prior studies~\citep{duquenne-etal-2023-speechmatrix, polyvoice, s2st-style}, each baseline model (Enc-Dec S2ST or Cascaded LMs) only supports a single translation direction and a separate model is needed for each direction. In contrast, our MSLM supports both directions in a single model.

Compared to ``Enc-Dec S2ST'' which has SOTA performance on this benchmark, our MSLM falls behind by a few points in ASR-BLEU. This is likely because MSLM employs causal LMs for unit translation which has less modeling capacity than the encoder-decoder architecture. However, this baseline relies on a unit-based vocoder for speech synthesis, which cannot transfer the speaker style. Hence, it achieves near zero speaker style similarity. Our MSLM utilizes the source speech as a style prompt for acoustic unit generation, which well preserves the speaker style and demonstrates a high similarity score in all cases.

Compared to ``Cascaded LMs'' which employs two separate AR LMs for semantic-to-semantic translation and semantic-to-acoustic generation, our proposed MSLM reduces the size of AR LM by a half but still achieves comparable or slightly better speaker style similarity and ASR-BLEU results in all cases. This verifies our hypothesis that the two conditional generation tasks do not interfere with each other. Instead, they might benefit each other by enhancing the modeling semantic units.

Additionally, we perform various ablation studies in Appendix~\ref{subsec:ablation} to demonstrate the effectiveness of our design choices.

\section{Conclusion}

We propose MSLM-S2ST, a multitask learning framework using speech LMs for speech-to-speech translation with style preservation. We achieve good translation quality and speaker style similarity without textual data or speaker style-aligned speech. 
Compared to existing LM studies on S2ST, MSLM enhances model parameter efficiency by (1) simplifying the cascaded semantic and acoustic LMs with a single model, (2) training with bidirectional translation to benefit from cross-lingual transfer. We outperform unidirectional cascaded LMs with only one-third of its parameters.

\section{Limitations and Risks}

\textbf{Limitations}. Although we did our best to
optimize for speech translation quality, toxic, biased, or false outputs produced by the model could remain.

\textbf{Ethical considerations}. The model is intended to support cross-lingual communication with speaker vocal style preserved. It might be misused for creating fake information, which is the unintended use of this model.
 
\bibliography{anthology,custom}

\newpage
\appendix

\section{Related Work}
\label{app:related-work}

\textbf{Speech language models}. Text LMs have successfully demonstrated in-context learning capabilities, paving the way for high-quality language understanding and generation. With recent advances in discrete audio representations, speech LMs have attracted an increasing research interest. Semantic units derived from SSL such as wav2vec 2.0~\citep{wav2vec2} and HuBERT~\citep{hubert} well capture the semantic information from speech. The GSLM family~\citep{lakhotia-etal-2021-generative, kharitonov-etal-2022-text, nguyen-etal-2023-generative} use semantic units to train and evaluate LMs on speech continuation. VoxtLM~\citep{voxtlm} integrates text tokens with semantic units to perform speech recognition, speech synthesis, text generation, and speech continuation in a single decoder-only LM. 

More recently, neural codec models such as EnCodec \cite{encodec} and SoundStream \cite{soundstream} generate multi-stream acoustic units which capture rich acoustic information. Acoustic units enable speech LMs to synthesize speech in a more fine-grained manner, e.g., style transfer in synthesis by the VALL-E family~\citep{valle, vallex}. Going beyond speech-only training, a line of research improves speech LMs by leveraging textual knowledge. AudioPaLM \cite{audiopalm} makes use of a pre-trained text LLM for initialization and aligned speech-text data for cross-modal training. 

\textbf{Speech-to-speech translation}. The study of S2ST starts from the preservation of semantic meaning from source to target speech. Existing approaches employ aligned speech as training data, and formulate S2ST as unit sequence prediction by converting target speech into semantic units \cite{textless-s2st}. Speech waveforms are then generated from units by a separately trained vocoder \cite{s2st-multi-target}. A major limitation of these conventional approaches is that the generated speech does not preserve acoustic information in the source speech such as the speaker's vocal style. Translatotron 2 \cite{translatotron2} supports vocal style transfer in translation, but it is trained on pseudo-labeled style-aligned speech and it utilizes speech-to-text data as auxiliary supervision to improve the translation accuracy.

\textbf{S2ST with LMs}.
More recent studies have explored speech LMs for vocal style-preserved S2ST, which are based on multiple cascaded LMs. PolyVoice \cite{polyvoice} is built upon three separate LMs. One is for semantic unit translation, one is for unit duration prediction, and the remaining one is for speech synthesis (i.e., acoustic unit generation) which is a variant of VALL-E~\cite{valle}. This design increases the size of the entire system and makes the training and deployment complicated.
Another concurrent work \cite{s2st-style} proposes different components, including speech-to-semantic-unit translation with flexible architectures and acoustic unit generation with an LM. It predicts only three acoustic streams and thus requires a separately trained unit-based vocoder for waveform generation.

Different from those approaches which employ separate models for semantic unit translation and acoustic unit generation~\cite{polyvoice, audiopalm, s2st-style}, our method adopts a single decoder-only LM for both tasks. It simplifies the overall architecture and reduces the total model size, which can be more efficient for training and deployment. Compared to prior studies which trains a separate system for each translation direction~\citep{polyvoice, s2st-style}, our model also supports bidirectional translation which even improves performance thanks to cross-lingual knowledge transfer.

\section{Proposed Method}
\label{app:method}

\begin{figure}[tb!]
    \centering
    \includegraphics[width=\linewidth]{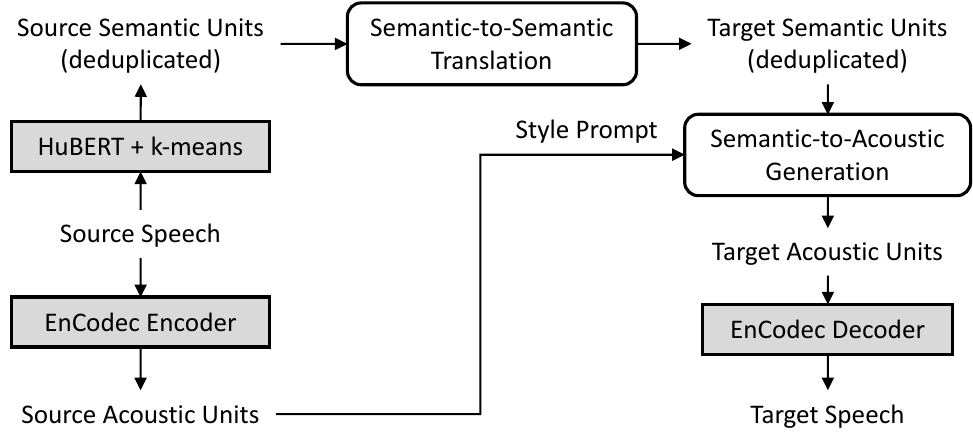}
    \caption{Overall pipeline of speaker style-preserved S2ST. The source speech is first translated to target semantic units and then converted to target acoustic units. Finally, the target speech is synthesized using a pre-trained EnCodec decoder.}
    \label{fig:overall}
\end{figure}

\autoref{fig:overall} illustrates the overall pipeline of speaker style-preserved S2ST.

\begingroup
\begin{table*}[htpb!]
    \centering
    \resizebox {\linewidth} {!} {
    \begin{tabular}{lcccccccc}
    \toprule
    \multirow{3}{*}{Model} & \multicolumn{4}{c}{Es-En} & \multicolumn{4}{c}{En-Es} \\
    & \multicolumn{2}{c}{EPST dev} & \multicolumn{2}{c}{EPST test} & \multicolumn{2}{c}{EPST dev} & \multicolumn{2}{c}{EPST test} \\
    & Spkr Sim. & BLEU & Spkr Sim. & BLEU & Spkr Sim. & BLEU & Spkr Sim. & BLEU \\
    \midrule
    MSLM & \textbf{0.400} & \textbf{24.78} & \textbf{0.393} & \textbf{21.50} & 0.430 & 21.41 & 0.421 & 22.09 \\
    ~~ w/o upsampling translation data & 0.375 & 23.74 & 0.369 & 21.42 & 0.416 & \textbf{21.64} & 0.406 & \textbf{22.46} \\
    ~~ unidirectional translation & 0.368 & 23.37 & 0.357 & 20.21 & 0.423 & 19.98 & 0.418 & 20.38 \\
    ~~ w/ OPT initialization & 0.391 & 24.50 & 0.380 & 21.17 & \textbf{0.437} & 21.00 & \textbf{0.429} & 20.67 \\
    \bottomrule
    \end{tabular}
}
\caption{Speaker style similarity ($\uparrow$) and ASR-BLEU ($\uparrow$) for ablation studies.
}
\label{tab:ablation}
\end{table*}
\endgroup

\section{Experiments}

\subsection{Training hyperparameters}
\label{sec:app_exp}

\paragraph{MSLM.} Our MSLM is trained with the Adam optimizer~\citep{adam-optimizer}. We set the learning rate as $2e-4$ for the AR component, and $5e-4$ for the NAR component. Both are trained for $200$k steps with a batch size of $4096$ tokens. We use $20$k warmup steps.

\paragraph{Cascaded LMs.} This baseline follows the same training hyperparameters as MSLM.

\paragraph{Enc-Dec S2ST.} The Enc-Dec model consists of a subsampler with two convolution layers, $12$ Transformer encoder layers and $12$ Transformer decoder layers. It has an auxiliary module of $6$-layer source unit decoder and is trained to predict source units as an additional task besides target semantic unit prediction. Enc-Dec S2ST predicts target semantic units, which is used to synthesize speech by a separately trained HiFi-GAN vocoder \cite{polyak21_interspeech}.
Enc-Dec models are trained for $150$k steps with a batch size of $4$k tokens. The learning rate is set as $5e-4$ for Es-En translation, and $3e-4$ for En-Es translation\footnote{We initially tried lr=$5e-4$ for En-Es model, and found that translation results were not good.}.

All models are periodically evaluated on the dev set, and the best checkpoint is saved for the final evaluation.

\subsection{Artifact License}
Here we discuss the license or terms of artifacts such as data, code and models used in this work. 

SpeechMatrix is publicly released under CC-BY-NC 4.0. EuroParl-ST is released under a Creative Commons license and is publicly accessible. The released data does not contain information that uniquely identifies individual people.

Speech tokenizers including HuBERT and EnCodec are released under MIT license. Baseline model Enc-Dec S2ST and vocoder are also under MIT license.

As for the evaluation tools, Whisper model used to transcribe generated speech is released under MIT license. WavLM used for speaker style similarity is under creative common public license.

\subsection{Ablation Studies}
\label{subsec:ablation}

We conduct the following ablation studies to verify the effectiveness of each design choice. Results are summarized in \autoref{tab:ablation}.

\textbf{Upsampling translation data in multitask learning.}
As discussed in Section~\ref{subsec:multitask-training}, when jointly learning semantic-to-semantic translation and semantic-to-acoustic generation, we find the translation task to be more challenging for the AR LM, so we upsample the translation data by 3 times. As shown in the second row in \autoref{tab:ablation}, upsamplig achieves comparable or better ASR-BLEU and improves speaker style similarity in all splits. Better semantic unit prediction yields better inputs to acoustic generation, which explains the enhanced quality of acoustic units and gains in speaker style similarity.

\textbf{Unidirectional vs bidirectional translation.}
The third row in \autoref{tab:ablation} shows that unidirectional MSLM clearly degrades both speaker style similarity and ASR-BLEU. This is likely because training MSLM with bidirectional translation tasks doubles the training data compared with a unidirectional model. Additionally, MSLM can benefit from the cross-lingual knowledge transfer.

\textbf{Initialization with a text LM.}
Prior studies suggest that initializing the AR LM with a pre-trained text LM can benefit certain speech generation tasks~\citep{twist, audiopalm}. We have experimented with initializing MSLM with a pre-trained OPT~\citep{opt} and found that although it speeds up convergence, the final performance does not improve or even slightly degrades, as shown in the last row in \autoref{tab:ablation}. Given the modality gap between speech and text, we conjecture that it needs more explorations into adapting text LM into speech.

\end{document}